\documentclass[11pt]{article}

\usepackage[final]{acl}

\usepackage{times}
\usepackage{latexsym}

\usepackage[T1]{fontenc}

\usepackage[utf8]{inputenc}

\usepackage{microtype}

\usepackage{inconsolata}

\usepackage{graphicx}

\usepackage{hyperref}
\usepackage{url}
\usepackage{subcaption}
\usepackage{soul}
\usepackage{hyperref}
\usepackage{amsmath}
\usepackage{amsfonts}
\usepackage{tcolorbox}
\usepackage{booktabs}
\usepackage{multicol}
\usepackage{multirow}
\usepackage{algorithm}
\usepackage{algpseudocode}
\usepackage{xcolor}
\usepackage{colortbl}
\title{Attention-Aligned Reasoning for Large Language Models}

\author{Hongxiang Zhang \and Yuan Tian \and Tianyi Zhang \\
        Purdue University \\ \texttt{hxxzhang@purdue.edu} \\ \texttt{tian211@purdue.edu} \\\texttt{tianyi@purdue.edu}}

\definecolor{lightblue}{RGB}{255, 255, 255}
\definecolor{lightgreen}{RGB}{230, 255, 233}
\definecolor{lightgray}{RGB}{225, 225, 225}
\definecolor{lightred}{rgb}{1.0, 0.8, 0.8}

\newcommand{\lightgreen}[1]{\sethlcolor{lightgreen}\hl{#1}}

\newcommand{\lightgray}[1]{#1}
\newcommand{\lightred}[1]{\sethlcolor{lightred}\hl{#1}}

\newcommand{\selfanchor}{\textsc{Atar}}
\newcommand{\headercolorlong}{\rowcolor{gray!15}}

\begin{document}
\maketitle

\begin{abstract}
Large Language Models (LLMs) tend to generate a long reasoning chain when solving complex tasks. 
However, as the reasoning chain extends, critical intermediate steps and the original prompt will be buried in the context, receiving insufficient attention and leading to errors. 
In this work, we present {\selfanchor}, a novel reasoning method that leverages the inherent reasoning structure to steer LLM attention. 
Our experiments show that {\selfanchor} outperforms SOTA methods across six benchmarks, achieving up to 15.39\% absolute improvement. 
Furthermore, with {\selfanchor}, ``non-reasoning'' models achieve comparable or even better performance compared to reasoning models of the same size in most benchmarks. 
Finally, our ablation studies show that the attention alignment component contributes significantly, and that these improvements are persist under different attention-steering backends.
\end{abstract}
\section{Introduction}

\begin{figure}[ht]
\centering
     \centering
     \includegraphics[width=1.\linewidth]{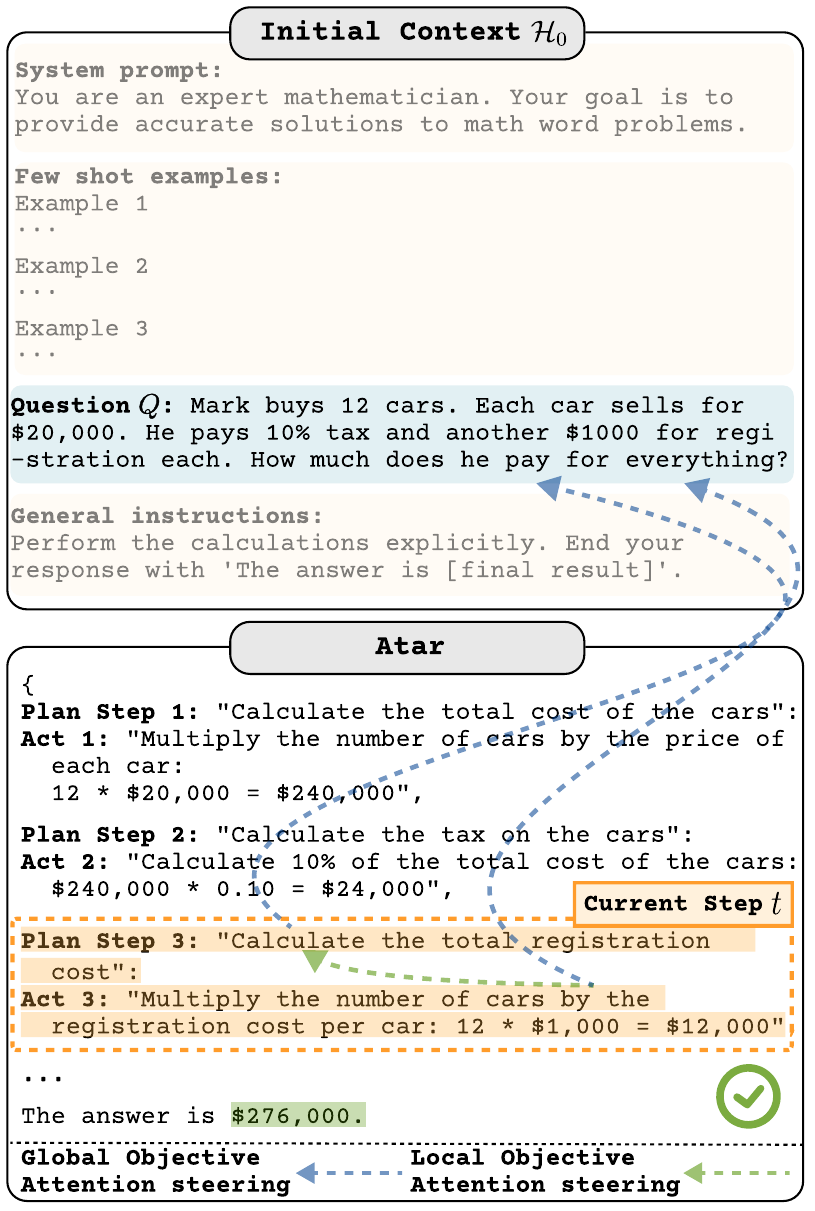}
        \caption{Overview of \selfanchor. By interleaving planning and action steps, our method explicitly steers attention to balance the global objective (the original question) and the local objective (the current plan step).}
    \vspace{-0.5cm}
    \label{fig:main}
\end{figure}

Solving complex tasks requires sophisticated reasoning capabilities. In practice, LLMs tend to generate long reasoning chains. However, as the context grows, the original prompt and key intermediate conclusions can be buried in the middle, receive insufficient attention, and consequently induce errors~\citep{liu-etal-2024-lost}. This degradation has been evidenced in many studies in the past~\citep{gu2024attention,chi2023attention,liu-etal-2024-lost,sun2024massive,yao2021self, tian2024selective, attention_decay, context_rot}. For example, relevant information placed in the middle of the prompt is often ignored and shows limited influence on generation~\citep{liu-etal-2024-lost}. More recently, researchers have identified an interesting phenomenon called ``attention sink''~\citep{xiao2023efficient}, where LLMs often allocate disproportionate attention to initial tokens. 
Excessive fixation on the static preamble can prevent the model from adequately attending to dynamic, task-specific constraints that emerge later in the reasoning process. 
Together, these effects may make models drift from the original intent and overlook critical intermediate constraints as the context grows.


Although attention steering methods~\citep{tian2024selective,zhang2023tell} have shown promise in steering attention toward salient content, they introduce their own limitations for multi-step reasoning. First, existing methods such as SPA~\citep{tian2024selective} and PASTA~\citep{zhang2023tell} rely on humans to select important tokens to pay attention to, which fails to generalize because the most relevant context shifts dynamically at each reasoning step. Second, these methods apply a fixed steering strength during generation. Such a static intervention can either be too weak to prevent drift or too strong to preserve fluent, correct reasoning.



To address these limitations, we propose {\selfanchor} (\underline{AT}tention-\underline{A}ligned \underline{R}easoning), a reasoning method that explicitly balances attention between the overarching goal and intermediate reasoning steps by leveraging the inherent structure of the reasoning chain. 
\autoref{fig:main} illustrates our method with an example 
where the current step provides the immediate sub-goal and the original user prompt specifies the global objective. As sub-goals evolve over the reasoning chain, the required degree of attention steering may vary at different steps. To accommodate this, we introduce a dynamic, step-level mechanism that adjusts the steering strength based on model confidence at each reasoning step.

We evaluate {\selfanchor} on six benchmarks and six base LLMs with varying sizes. 
Compared with SOTA reasoning methods such as Tree-of-Thought~\citep{yao2023tree} and Re-Reading~\cite{xu2024re}, {\selfanchor} consistently achieves better performance by up to 15.39\% absolute improvement. Furthermore, compared with reasoning models such as Qwen3-thinking~\citep{qwen3technicalreport} and Phi-4-reasoning~\citep{abdin2025phi}, {\selfanchor} helps the corresponding non-reasoning models in the same model family achieve better or comparable performance in most settings. Our ablation studies show that the attention alignment component contributes significantly, with improvements persisting across different attention-steering backends.

\section{Method}

\begin{figure}
\centering
     \centering
     \includegraphics[width=1.\linewidth]{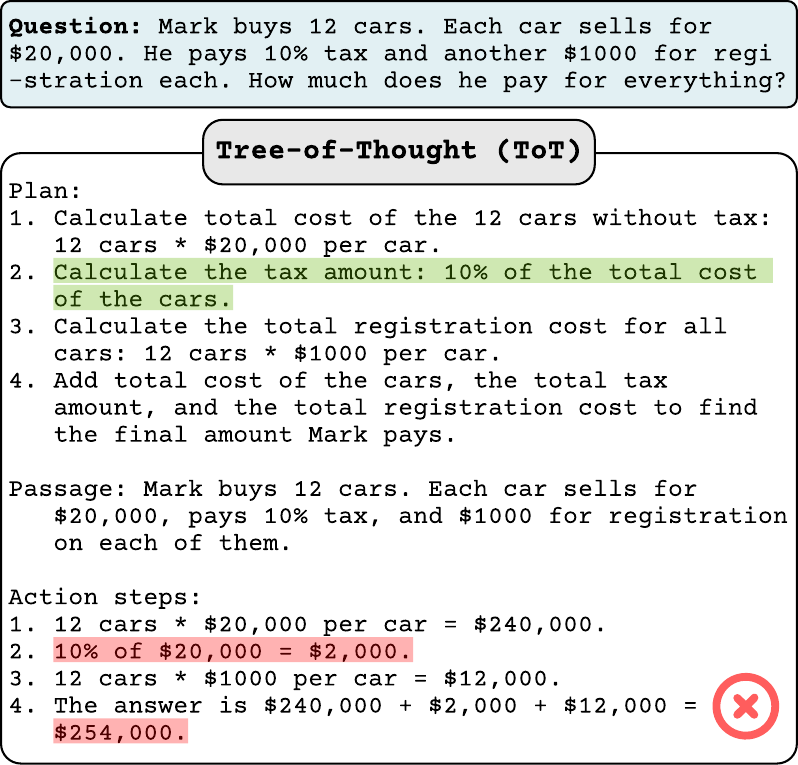}
        \caption{An example in Tree-of-Thought. ToT successfully generates a correct plan. The corresponding action is separated from the plan by a large amount of intervening context, thus deviating from the plan's instruction.}
    \vspace{-0.6cm}
    \label{fig:tot}
\end{figure}

\paragraph{Attention Aligned Reasoning.}\label{Automatic Anchoring Strategy}
In complex reasoning tasks, the input context typically includes diverse information components, such as the system prompt, few-shot examples, relevant knowledge, and general instructions. Within this context, the core question $Q$, is often embedded in the middle. For simplicity, we denote $Q$ as the original question and the contextual information excluding $Q$ as the preamble variable $\Phi$. The initial context is thus defined as $\mathcal{H}_0 = \{\Phi, Q\}$. In our method, we posit $Q$ as the global objective that should consistently guide the generation process. 

Prior research has demonstrated the effectiveness of the ``Plan-then-Solve'' paradigm for complex reasoning, in which a model generates a comprehensive plan before executing step-by-step actions~\citep{zhou2022least,jiang2024self,zhou2024self,wang2023plan}. However, when executing later actions, the corresponding plan steps may be separated by a large amount of intervening context, interfering the generation. As illustrated in Figure~\ref{fig:tot}, as the model generates actions, the corresponding plan step may be obscured by the accumulation of intermediate context, leading to attention drift. Inspired by findings that LLMs often attend to immediately generated tokens~\citep{liu-etal-2024-lost}, we place each action immediately following the corresponding plan step. Thus, we formalize the reasoning process as a trajectory $\tau$ comprising $T$ steps, where each step $t$ interleaves a planning step $s_t$ and an action step~$a_t$. The complete trajectory is represented as $\tau = \{s_1, a_1, \dots, s_T, a_T\}$.

Within this structure, we found that plans offer distinct reference points that can naturally serve to guide attention alignment. Each plan step $s_t$ explicitly specifies the sub-goal to be accomplished at step $t$. 
Therefore, we interpret the current plan step $s_t$ as the {\em local objective}, while the original query $Q$ defines the {\em global objective}. 
To generate the action $a_t$, {\selfanchor} steers the model's attention to both the global objective $Q$ and the local objective $s_t$. This allows the model to balance attention between the overarching goal and intermediate reasoning steps:
\begin{equation}
a_t = f(\mathcal{H}_{0}\oplus\mathcal{\tau}_{t-1} \oplus s_t, \{Q, s_t\}).
\end{equation}
where we denote the generation function with attention steering as $f(x, \mathcal{S})$, where $x$ represents the sequence of preceding tokens (i.e., the context) and $\mathcal{S}$ represents the specific set of tokens to which the model's attention is steered. The $\oplus$ indicates string concatenation.

Similarly, generating the next plan step $s_t$ may be interfered by the growing context, particularly from previously generated actions. To ensure that new sub-goals remain faithful to the original question, \selfanchor\ steers the generation of each plan step toward the global objective $Q$:
\begin{equation}
s_t = f(\mathcal{H}_{0}\oplus\mathcal{\tau}_{t-1}, \{Q\}).
\end{equation}

\paragraph{Dynamic Attention Alignment.}
    As reasoning trajectories progress, the required degree of attention alignment may vary at different steps. 
    Prior work has explored using an LLM's predicted probability distribution as an approximated signal to assess prediction quality~\citep{confidence1, confidence2, jiang-etal-2023-active, spiess2024calibration, kadavath2022language, orgad2024llms}. 
    High token-level confidence generally corresponds to locally stable reasoning~\citep{varshney2022can,jiang2021can}, where the model’s internal trajectory is already coherent and requires only weak steering. Low confidence, on the other hand, is often correlated with uncertainty or inconsistency. In such cases, stronger steering helps realign the model toward the relevant context.\footnote{While token-level confidence is not universally reliable and can be mis-calibrated, we adopt it as a lightweight, approximate indicator of local stability.}

    Therefore, we introduce a step-level steering strength $\omega_j$ that is dynamically adjusted based on model confidence.\footnote{Our strength adjustment strategy builds upon the confidence-modulated strength strategy in SPA~\citep{tian2024selective}. While SPA adjusts the strength based on confidence for each logit at the vocabulary level, we introduce an additional factor to adjust the strength for each step.} Let $P_j = \{p_{j,1}, p_{j,2}, \ldots, p_{j,m} \}$ denote the probabilities of the $m$ tokens generated at step $j$. We define the step confidence as the harmonic mean of probabilities:
    \begin{equation}
        \mathrm{conf}(P_j) = \frac{m}{ \sum_{k=1}^{m} \frac{1}{p_{j,k}} }
    \end{equation}

    We adopt the harmonic mean because it is sensitive to low-probability tokens~\citep{kai-etal-2024-sh2, krishnan2024enhancing}, so a few uncertain predictions can sharply lower the overall score.
    This behavior is well-suited to multi-step reasoning, where attention drift is often associated with locally uncertain generations~\citep{kai-etal-2024-sh2}. By comparison, alternatives such as entropy or the arithmetic mean smooth uncertainty more evenly across tokens and thus may fail to emphasize the low-confidence tail, where intervention is most critical. We experimented with alternative designs in Appendix~\ref{appendix:mean} and demonstrated that harmonic mean achieves the best performance.
        
    We then calculate the dynamic strength $\omega_j$ inversely with this confidence:
    
    \begin{equation}
        \omega_j = \frac{\omega \cdot \sum_{k=1}^{m} \frac{1}{p_{j,k}} }{m},
    \end{equation}
    where $\omega$ is the base steering strength and $\omega_j$\footnote{We clip $\omega_j$ to the range recommended by SPA~\citep{tian2024selective}. Small probabilities can increase steering, as intended, but cannot produce unbounded numerical values.} is the steering strength at decoding step $j$. Consequently, attention steering is amplified during highly uncertain decoding steps and attenuated when the model is confident.
    
    To implement the actual attention intervention, we adopt Selective Prompt Anchoring (SPA)~\citep{tian2024selective} as a lightweight steering backend. We experimented with an alternative attention steering method, PASTA~\citep{zhang2023tell}, and demonstrated that our method achieves persistent performance with the alternative steering method in Sec.~\ref {sec: alternative backbones}.
\section{Experiments}

\subsection{Benchmarks}
\label{sec:benchmarks}
We evaluated {\selfanchor} on six benchmarks. The first three benchmarks are about math reasoning, including GSM8K~\citep{cobbe2021training}, AQuA~\citep{ling2017program}, and MATH-500~\citep{lightman2023lets}. These three benchmarks span increasing difficulty levels, from grade-school math problems~\citep{cobbe2021training} to GMET/GRE \citep{ling2017program} level and competition-level problems \citep{hendrycks2021measuring}. The next two benchmarks are about commonsense reasoning, including StrategyQA~\citep{geva2021did} and Thinking for Doing (T4D)~\citep{zhou2023far}. 
Finally, following \citet{zou2025many,juneja2025task,wei2022chain}, we select challenging tasks in BIG-Bench Hard (BBH)~\citep{suzgun2022challenging} as the last benchmark.
Specifically, this benchmark covers six types of reasoning tasks,  including {\em Disambiguation QA} (DQA), {\em Causal Judgment} (CJ), {\em Date Understanding} (DU), {\em Logical Deduction} (LD), {\em Salient Translation Error Detection} (STED), and {\em Snarks}. We report final-answer accuracy across all benchmarks.\footnote{Prompt templates and evaluation details are provided in \autoref{appendix: Implementation and evaluation details}.}

\subsection{Models and Baselines}
\textbf{Base Models.} We experiment with six non-reasoning LLMs spanning various scales as the backbone of {\selfanchor}. 
We select Llama-3.1-8B-Instruct~\citep{grattafiori2024llama}, Llama-3.2-3B-Instruct~\citep{grattafiori2024llama}, Phi-4-mini~\citep{abouelenin2025phi}, Qwen3-4B-Instruct-2507~\citep{qwen3technicalreport}, Phi-4~\citep{abdin2024phi}, and Qwen3-30B-A3B-Instruct-2507~\citep{qwen3technicalreport}.

\textbf{Comparison Baselines.}
We compare our method against four representative reasoning methods for LLM reasoning.
First, we include \textbf{Chain-of-Thought (CoT)}~\citep{wei2022chain,kojima2022large}, a widely used baseline for LLM reasoning. 
Second, we include \textbf{Plan-and-Solve+ (PS+)}~\citep{wang2023plan}, which first generates a plan and then solves the problem. 
Third, we include \textbf{Re-Reading (RE2)}~\citep{xu2024re}, which asks the model to read the question again and then solve the problem. Lastly, we include \textbf{Tree-of-Thought (ToT)}~\citep{yao2023tree}, which explores multiple reasoning paths through deliberate tree search.

\begin{table*}[t]
    \scriptsize
    \centering
    \setlength{\tabcolsep}{4pt}
    \caption{Evaluation results on six benchmarks. Best results are shown in \textbf{bold}, and those indicating a performance gain compared to Chain-of-Thought are shown in \lightgreen{green}.}
    \resizebox{\textwidth}{!}{
        \begin{tabular}{ll|ccc|cc|c}
    \toprule
    \multirow{2.5}{*}{\textbf{Model}} & \multirow{2.5}{*}{\textbf{Method}} & \multicolumn{3}{c|}{Math} & \multicolumn{2}{c|}{CommonSense} &  \multirow{2.5}{*}{BBH} \\
    \cmidrule(lr){3-5} \cmidrule(lr){6-7}
    & & GSM8K & AQuA & MATH-500 & StrQA & T4D \\
    \midrule
    \multirow{5}{*}{Llama3.2-3B} & CoT & 73.16 & 46.06 & 40.80 & 66.64 & 32.98 & 35.09 \\
        & PS+ & \lightgreen{76.95 (+3.79)} & \lightgray{38.58 (-7.48)} & \lightgreen{42.20 (+1.40)} & \lightgray{62.79 (-3.85)} & \lightgray{31.81 (-1.17)} & \lightgreen{40.51 (+5.42)} \\
        & RE2 & \lightgreen{74.68 (+1.52)} & \lightgreen{47.28 (+1.22)} & \lightgreen{41.00 (+0.20)} & \lightgray{65.03 (-1.61)} & \lightgray{30.32 (-2.66)} & \lightgreen{38.02 (+2.93)} \\
        & ToT & \lightgray{70.89 (-2.27)}& \lightgray{35.04 (-11.02)}& \lightgray{32.80 (-8.00)} & \lightgray{60.44 (-6.20)} & \lightgreen{40.80 (+7.82)} & \lightgreen{41.32 (+6.23)}\\
        & \textbf{{\selfanchor}} & \lightgreen{\textbf{78.32 (+5.16)}} & \lightgreen{\textbf{48.43 (+2.37)}} & \lightgreen{\textbf{42.40 (+1.60)}} & \lightgreen{\textbf{67.55 (+0.91)}} & \lightgreen{\textbf{43.79 (+10.81)}} & \lightgreen{\textbf{50.48 (+15.39)}} \\
    \midrule
    \multirow{5}{*}{Phi-4-mini} & CoT & 84.53 & 61.81 & 61.20 & 67.03  & 39.54 & 60.51\\
        & PS+ & \lightgreen{88.01 (+3.39)} & \lightgreen{62.02 (+0.21)} & \lightgreen{62.40 (+1.20)} & \lightgray{58.86 (-8.17)} & \lightgreen{41.16 (+1.62)} & \lightgray{59.63 (-0.88)} \\
        & RE2& \lightgreen{87.33 (+2.80)} & \lightgray{59.06 (-2.75)} & \lightgray{60.80 (-0.40)} & \lightgray{61.83 (-5.20)} & \lightgreen{44.50 (+4.96)} & \lightgreen{61.39 (+0.88)} \\
        & ToT & \lightgreen{85.22 (+0.69)} & \lightgreen{66.93 (+5.12)} & 57.60 (-3.60) & \lightgray{61.22 (-5.81)} & \lightgreen{48.05 (+8.51)} & \lightgray{51.72 (-8.79)}\\
        & \textbf{{\selfanchor}} & \lightgreen{\textbf{88.02 (+3.49)}} & \lightgreen{\textbf{68.50 (+6.69)}} & \lightgreen{\textbf{62.60 (+1.40)}} & \lightgreen{\textbf{68.69 (+1.66)}} & \lightgreen{\textbf{49.47 (+9.93)}} & \lightgreen{\textbf{62.42 (+1.91)}} \\
    \midrule
    \multirow{5}{*}{Qwen3-4B} & CoT & 92.27& 73.62 & 80.20 & 68.03  & 70.21 & 73.33\\
        & PS+ & \lightgray{91.43 (-0.84)} & \lightgray{71.65 (-1.97)} & \lightgreen{82.40 (+2.20)} & \lightgreen{68.56 (+0.53)} & \lightgray{59.82 (-10.39)} & \lightgray{70.99 (-2.34)} \\
        & RE2& \lightgray{88.70 (-3.57)} & \lightgreen{75.98 (+2.36)} & \lightgreen{81.00 (+0.80)} & \lightgreen{69.83 (+1.80)} & \lightgray{67.02 (-3.19)} & \lightgreen{\textbf{75.75 (+2.42)}} \\
        & ToT & \lightgray{89.79 (-2.48) }& \lightgreen{74.80 (+1.18)} & \lightgreen{86.60 (+6.40)} & \lightgreen{69.89 (+1.86)} & \lightgray{52.24 (-17.97)} & \lightgray{69.82 (-3.51)}\\
        & \textbf{{\selfanchor}} & \lightgreen{\textbf{93.56 (+1.29)}} & \lightgreen{\textbf{79.92 (+6.30)}} & \lightgreen{\textbf{88.60 (+8.40)}} & \lightgreen{\textbf{70.13 (+2.10)}} & \lightgreen{\textbf{71.56 (+1.35)}} & \lightgreen{75.31 (+1.98)} \\        
    \midrule
    \multirow{5}{*}{Llama3.1-8B} & CoT & 77.03 & 50.79 & 43.40 & 70.24 & 26.77 & 49.45 \\
        & PS+ & \lightgreen{77.18 (+0.15)} & \lightgray{48.65 (-2.14)} & \lightgray{40.60 (-2.80)} & \lightgray{65.85 (-4.39)} & \lightgreen{28.79 (+2.02)} & \lightgreen{51.72 (+2.27)} \\
        & RE2 & \lightgray{72.10 (-4.93)} & \lightgreen{51.97 (+1.18)} & \lightgreen{46.20 (+2.80)} & \lightgray{69.74 (-0.50)} & \lightgreen{28.55 (+1.78)} & \lightgreen{54.58 (+5.13)} \\
        & ToT & \lightgray{70.28 (-6.75)} & \lightgreen{52.76 (+1.97)} & \lightgreen{43.60 (+0.20)} & \lightgray{69.58 (-0.66)} & \lightgreen{35.75 (+8.98)} & \lightgreen{50.47 (+1.02)}\\
        & \textbf{{\selfanchor}} & \lightgreen{\textbf{83.93 (+6.90)}} & \lightgreen{\textbf{55.51 (+4.72)}} & \lightgreen{\textbf{50.20 (+6.80)}} & \lightgreen{\textbf{73.54 (+3.30)}} & \lightgreen{\textbf{40.01 (+13.24)}} & \lightgreen{\textbf{58.53 (+9.08)}} \\
    \midrule
    \multirow{5}{*}{Phi-4} & CoT & 89.08 & 68.11 & 74.60 & 77.51 & 73.94 & 72.08\\
        & PS+ & 91.36 (+2.28) & \lightgreen{72.75 (+4.64)} & \lightgreen{76.40 (+2.00)} & \lightgray{75.59 (-1.92)} & \lightgreen{76.24 (+2.30)} & \lightgray{68.42 (-3.66)} \\
        & RE2 & \lightgreen{89.84 (+0.76)} & \lightgreen{69.29 (+1.18)} & \lightgreen{74.80 (+0.20)} & \lightgray{76.90 (-0.61)} & \lightgreen{75.79 (+1.85)} & \lightgreen{74.94 (+2.86)} \\
        & ToT & \lightgreen{90.60 (+1.52)} & \lightgreen{75.67 (+7.56)} & \lightgreen{79.40 (+4.80)} & \lightgray{70.02 (-7.49)} & \lightgreen{83.95 (+10.01)} & \lightgray{64.10 (-8.70)} \\
        & \textbf{{\selfanchor}} & \lightgreen{\textbf{94.38 (+5.30)}} & \lightgreen{\textbf{79.13 (+11.02)}} & \lightgreen{\textbf{81.00 (+6.40)}} & \lightgreen{\textbf{77.82 (+0.31)}} & \lightgreen{\textbf{85.99 (+12.05)}} & \lightgreen{\textbf{75.31 (+3.23)}} \\
    \midrule
    \multirow{5}{*}{Qwen3-30B} & CoT & 88.07& 81.10 & 89.40 & 78.60 & 84.92 & 72.80\\
        & PS+ & \lightgreen{89.04 (+0.97)} & \lightgray{80.83 (-0.27)} & \lightgreen{91.20 (+1.80)} & \lightgreen{78.70 (+0.10)} & \lightgray{82.62 (-2.30)} & \lightgray{71.79 (-1.01)}\\
        & RE2 & \lightgreen{90.54 (+2.47)} & \lightgreen{82.28 (+1.18)} & \lightgreen{89.80 (+0.20)} & \lightgreen{78.91 (+0.31)} & \lightgreen{\textbf{89.36 (+4.44)}} & \lightgreen{\textbf{75.68 (+2.88)}}\\
        & ToT & \lightgreen{90.43 (+2.36)} & \lightgreen{\textbf{87.06 (+5.96)}} & \lightgreen{91.60 (+2.20)} & \lightgray{77.41 (-1.19)} & \lightgray{83.97 (-0.95)} & \lightgreen{73.55 (+0.75)} \\
        & \textbf{{\selfanchor}} & \lightgreen{\textbf{92.16(+4.09)}} & \lightgreen{83.46 (+2.36)} & \lightgreen{\textbf{93.60 (+4.20)}} & \lightgreen{\textbf{79.65 (+1.05)}} & \lightgreen{85.56 (+0.64)} & \lightgreen{74.43 (+1.63)}\\
        \bottomrule
        \end{tabular}
    }
    \vspace{-0.4cm}
    \label{tab:main}
\end{table*}

\begin{table}[ht]
    \scriptsize
    \centering
    \setlength{\tabcolsep}{2pt}
    \caption{Performance of {\selfanchor} across BBH task categories. Full task names are provided in Sec. \ref{sec:benchmarks}.}
    \resizebox{\linewidth}{!}{
    \begin{tabular}{lcccccc}
    \toprule
    \multirow{2.5}{*}{\textbf{Model}} &  \multicolumn{6}{c}{\textbf{BBH}} \\
    \cmidrule(lr){2-7}
    &
    \textbf{DQA} & \textbf{CJ}  & \textbf{DU} & \textbf{LD} & \textbf{STED} & {\textbf{Snarks}}\\
    \midrule
        Llama3.2-3B & 52.10 & 54.01 & 52.40 & 40.40 & 38.40 & 74.47\\
        \midrule
        Phi4-mini & 60.80 & 62.03 & 69.60 & 51.20 & 56.00 & 79.78\\
        \midrule
        Qwen3-4B & 74.00 & 61.50 & 77.20 & 89.60 & 64.40 & 84.27 \\
        \midrule
        Llama3.1-8B & 60.80 & 53.48 & 64.00 & 51.20 & 52.80 & 71.35\\
        \midrule
        Phi-4 & 70.40 & 65.24 & 87.20 & 81.20 & 59.20 & 90.45 \\
        \midrule
        Qwen3-30B & 72.00 & 65.78 & 78.40 & 80.00 & 64.00 & 88.20\\
    \bottomrule
    \end{tabular}
    }
    \label{tab: bbh}
    \vspace{-0.8cm}
\end{table}

Furthermore, we consider four reasoning LLMs that have a comparable size to the non-reasoning base models used in {\selfanchor} as baselines to investigate whether non-reasoning LLMs combined with \selfanchor\ achieve competitive performance against reasoning models across different model scales, including Phi-4-mini-reasoning~\citep{abouelenin2025phi}, Qwen3-4B-Thinking-2507~\citep{qwen3technicalreport}, Phi-4-reasoning~\citep{abdin2024phi}, and Qwen3-30B-A3B-Thinking-2507~\citep{qwen3technicalreport}.


\subsection{Main Results}
\label{main result}
\paragraph{Math Reasoning.}
    Math reasoning represents one of the most challenging aspects of LLM reasoning capabilities. As shown in \autoref{tab:main}, {\selfanchor} consistently improves significant accuracy across three math benchmarks. 
    Specifically, it achieves absolute improvements of around 5\% on GSM8K, AQuA and up to 8.4\% on MATH-500 across most models.

     Although PS+, RE2, and ToT demonstrate some effectiveness in improving LLM's math reasoning capabilities, \autoref{tab:main} shows that such improvement is not consistent. For some benchmarks and base models, these methods achieve even worse performance compared with CoT.  Specifically, they tend to be more effective on larger LLMs (Phi-4-15B and Qwen3-30B). This may be because larger models are more capable of following instructions and external guidance to align reasoning trajectories.
    
     
    As mentioned in Section~\ref{sec:benchmarks}, GSM8K, AQuA, and MATH-500 represent three levels of math reasoning difficulty. \autoref{tab:main} shows that {\selfanchor} consistently improves the performance of different LLMs across all three benchmarks, while the other three methods show different levels of performance degradation compared to CoT on certain models and benchmarks.  This highlights the robustness and generalization of {\selfanchor} across diverse model architectures and reasoning difficulty levels.

\paragraph{Commonsense Reasoning.}
    StrategyQA requires multi-hop reasoning over commonsense knowledge. As shown in \autoref{tab:main}, {\selfanchor} persistently improves accuracy across six base LLMs. In contrast, PS+, RE2, and ToT occasionally outperform the baseline CoT method.

    T4D is a grounded Theory of Mind reasoning task~\citep{baron1985does, frith2003development}, which requires the model to predict reasonable actions based on a character's hidden thoughts and perspective. 
    PS+, RE2, and ToT exhibit mixed performance on T4D. They tend to be more effective in larger models. In comparison, {\selfanchor} demonstrates significant performance gains of over 9\% on four of the six LLMs.
    
\paragraph{BBH.}
    BBH aggregates challenging algorithmic and symbolic tasks. {\selfanchor} demonstrates average performance gains ranging from 1.61\% to 15.39\%\footnote{We detail the subtask performance in \autoref{appendix:bbh}}. 
    When analyzing specific task categories in \autoref{tab: bbh}, we find that those requiring tracking of intermediate reasoning benefit the most, for example \textit{date understanding}, and \textit{logical deduction}. We attribute this to {\selfanchor}'s attention steering that augments critical reasoning steps and the original question throughout generation. In contrast, PS+, RE2, and ToT show inconsistent improvements.

\paragraph{Comparison to RL-enhanced Reasoning Models}
Recent advances in reasoning capabilities have been dominated by reinforcement learning-enhanced ``reasoning'' models that employ extensive internal reasoning chains during inference. However, these models are costly to fine-tune and require large-scale training data. This raises a question: \textit{Can non-reasoning LLMs combined with {\selfanchor} achieve competitive performance against reasoning models?}

\begin{table*}[t]
    \scriptsize
    \centering
    \setlength{\tabcolsep}{4pt}
    \caption{Evaluation comparison with reasoning models within the same family and size.}
    \resizebox{0.75\textwidth}{!}{
        \begin{tabular}{l|ccc|cc|c}
    \toprule
    \multirow{2.5}{*}{\textbf{Model}} & \multicolumn{3}{c|}{Math} & \multicolumn{2}{c|}{CommonSense} &  \multirow{2.5}{*}{BBH} \\
    \cmidrule(lr){2-4} \cmidrule(lr){5-6}
    & GSM8K & AQuA & MATH-500 & StrQA & T4D \\
    \midrule
    Phi-4-mini-reasoning  & 85.37 & 67.97 & \textbf{87.80} & 66.38 & 45.04 & 59.85 \\
    Phi-4-mini (w/ \textbf{{\selfanchor}}) & \textbf{88.02} &  \textbf{68.50} & 62.60 &  \textbf{68.69} &  \textbf{49.47} & \textbf{62.42} \\
    \midrule
    Qwen3-4B-Thinking  & 92.27 & 67.32 & \textbf{93.20} & 68.31 & \textbf{73.40} & \textbf{75.34} \\
    Qwen3-4B (w/ \textbf{{\selfanchor}}) & \textbf{93.56} & \textbf{79.92} & 88.60 & \textbf{70.13} & 71.56 & 75.31\\
    \midrule
    Phi-4-reasoning & 91.20 & \textbf{83.20} &  \textbf{91.20}& 77.61 & 74.11 & 74.98 \\
    Phi-4 (w/ \textbf{{\selfanchor}}) & \textbf{94.38} & 79.13 & 81.00 & \textbf{77.82} & \textbf{85.99} & \textbf{75.31} \\
    \midrule
    Qwen3-30B-A3B-Thinking & \textbf{93.75} & 83.26 & \textbf{95.60}& 77.26 & 80.96 &  \textbf{76.54} \\
    Qwen3-30B (w/ \textbf{{\selfanchor}}) & 92.16 & \textbf{83.46} & 93.60 & \textbf{79.65} & \textbf{85.56} & 74.43 \\
        \bottomrule
        \end{tabular}
    }
    \label{tab:thinking_comparison}
    \vspace{-0.2cm}
\end{table*}

\begin{table}[ht]
    \centering
    \setlength{\tabcolsep}{4pt}
    \caption{Performance of {\selfanchor} under different backends, including SPA~\citep{tian2024selective}, and PASTA~\citep{zhang2023tell}.}
    \resizebox{\linewidth}{!}{
    \scriptsize
    \begin{tabular}{l|ccc}
    \toprule
    \textbf{Method} & \textbf{Math} & \textbf{CommonSense} & \textbf{BBH} \\
    \midrule
    CoT & 24.17 & 57.59 & 31.67 \\
    PS+ & 23.66 & 55.02 & 32.89\\
    RE2 & 24.96 & 58.38 & 33.60 \\
    ToT & 22.56 & 49.46 & 28.74\\
    \midrule
    \textbf{{\selfanchor} (w/ SPA)} & \textbf{29.17} & 59.77 & \textbf{35.01} \\
    \textbf{{\selfanchor} (w/ PASTA)} & 26.53 & \textbf{60.43} & 33.71 \\
    \bottomrule
    \end{tabular}
    }
    \label{tab: pasta}
    \vspace{-0.5cm}
\end{table}

To investigate this question, we compare our method applied to non-reasoning LLMs against the corresponding reasoning models version. As illustrated in Table~\ref{tab:thinking_comparison}, applying {\selfanchor} to non-reasoning models achieves competitive or superior performance compared to RL-enhanced reasoning models on most benchmarks and models. Specifically, {\selfanchor} closes the performance gap significantly in three math benchmarks; On commonsense reasoning tasks and BBH, \selfanchor\ exceeds reasoning model performance on most benchmarks and LLMs.

Notably, reasoning models show advantages on tasks where the non-reasoning models have weak baseline performance. Specifically, Phi-4, Qwen3-4B, and Phi-4-mini show large gaps on MATH-500. We hypothesize that this pattern is largely attributable to reinforcement learning in post-training. RL is particularly effective for verifiable reasoning tasks such as mathematics, where correctness can be directly evaluated and used as a reliable learning signal~\citep{qwen3technicalreport,deepseekai2025deepseekr1incentivizingreasoningcapability}. In contrast, for tasks where non-reasoning models already demonstrate strong performance, reinforcement learning provides limited improvement. This pattern is also observed in \citet{kirk2023understanding}. Nevertheless, \selfanchor\ shows consistent performance improvements across all tasks and difficulty levels.

In summary, instead of learning implicit reasoning patterns through expensive training, {\selfanchor} leverages the explicit reasoning structure for attention alignment to improve the reasoning capabilities. These findings suggest that {\selfanchor} can serve as an effective alternative to computationally expensive RL-enhanced reasoning.

\subsection{Generalizability to Attention Steering Backbones}
\label{sec: alternative backbones}
{\selfanchor} is orthogonal to the choice of attention steering method. To validate this, we experiment {\selfanchor} with an alternative steering backend, PASTA~\citep{zhang2023tell}. In this experiment, we follow the original PASTA setup with Llama~\citep{touvron2023llama}, since PASTA only supports limited backbones and defaults to Llama for evaluation. Accordingly, we evaluate {\selfanchor} with PASTA across three representative benchmarks from our six primary evaluation benchmarks: AQuA, StrQA, and BBH. 

As shown in \autoref{tab: pasta}, {\selfanchor} consistently improves performance over other reasoning methods under different attention steering backends. These results suggest that {\selfanchor} generalizes across attention steering backbones.

\subsection{Ablation Study}
\label{sec: ablation attention steering}
\textbf{Attention Steering.}
To isolate the contribution of the attention steering, we compare {\selfanchor} against a variant without attention steering. Additional steering strategies and case studies are provided in \autoref{appendix: alternative design} and \autoref{appendix:case}.

As shown in \autoref{tab:ablation structure}, {\selfanchor} consistently outperformed its variant without attention steering across all benchmarks. This ablation confirms that planning and structured reasoning alone are insufficient to prevent attention drift.

\textbf{Dynamic Attention Steering.}
We further compare {\selfanchor} against an ablated version that only performs static attention steering without dynamic token selection and strength tuning. Following the main experiments, we still use SPA~\citep{tian2024selective} as the attention steering backbone. Note that without dynamic token selection, we need to manually select which tokens to steer attention to during the reasoning and generation process, which does not scale to a large experiment. Therefore, we follow the original setup from SPA, which steers to the original user question for simplicity and automation. 

As shown in \autoref{tab:ablation structure}, {\selfanchor} consistently outperformed its variant without the dynamic mechanism across all benchmarks. These results indicate that simply combining planning with attention steering is insufficient. The gains of {\selfanchor} stem from dynamic attention steering, which maintains focus as reasoning evolves.

\begin{table*}
    \scriptsize
    \centering
    \setlength{\tabcolsep}{3pt}
    \caption{Ablation study comparing {\selfanchor} with and without attention steering. Results show that attention steering consistently improves performance. Best results are highlighted in \textbf{bold}.}
    \resizebox{0.92\textwidth}{!}{
        \begin{tabular}{ll|ccc|cc|c}
    \toprule
    \multirow{2.5}{*}{\textbf{Model}} & \multirow{2.5}{*}{\textbf{Method}} & \multicolumn{3}{c|}{Math} & \multicolumn{2}{c|}{CommonSense} &  \multirow{2.5}{*}{BBH} \\
    \cmidrule(lr){3-5} \cmidrule(lr){6-7}
    & & GSM8K & AQuA & MATH-500 & StrQA & T4D \\
    \midrule
    \multirow{2}{*}{Llama3.1-8B} 
    & {\selfanchor} (w/o attention steering) & 79.01 & 53.15 & 47.60 & 71.17 & 35.28 & 52.77\\
    & {\selfanchor} (w/o dynamic mechanism) & 80.13 & 48.03 & 48.80 & 73.18 & 38.17 & 53.64\\
    & {\selfanchor} & \textbf{83.93} & \textbf{55.51} & \textbf{50.20} & \textbf{73.54} & \textbf{40.01} & \textbf{58.53}  \\
    
    \midrule
    \multirow{2}{*}{Phi-4-mini-4B} 
    & {\selfanchor} (w/o attention steering) & 78.77 & 58.66 &55.40 & 68.60 & 38.83 & 57.44 \\
    & {\selfanchor} (w/o dynamic mechanism) & 84.32  & 62.60 & 57.20 & 67.97 & 48.47 & 60.80 \\
    & {\selfanchor} & \textbf{88.02} & \textbf{68.50} & \textbf{62.60} & \textbf{68.69} & \textbf{49.47} & \textbf{62.42} \\
    
    \bottomrule
    \end{tabular}
    }
    \label{tab:ablation structure}
    \vspace{-0.3cm}
\end{table*}

\begin{figure}[ht]
    \centering
    \begin{subfigure}[t]{0.48\textwidth}
        \raggedright
        \includegraphics[width=\textwidth]{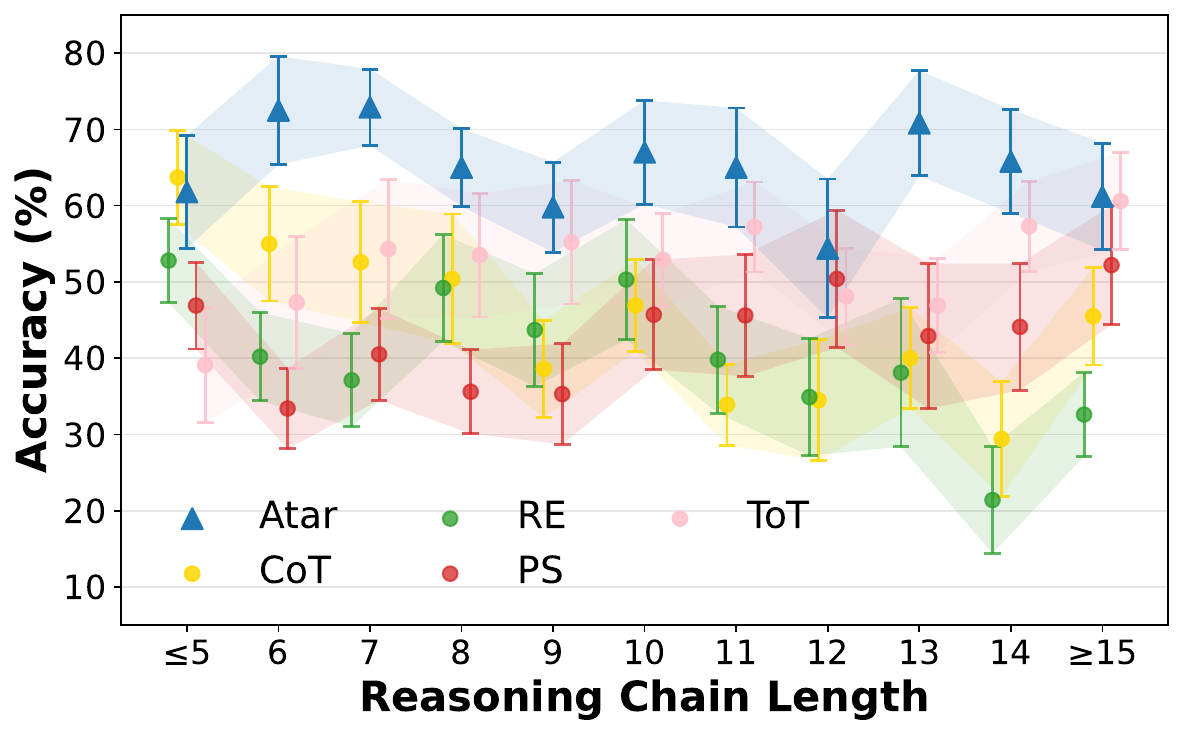}
        \caption{Reasoning chain length and accuracy comparison with 95\% confidence intervals. }
        \label{fig:combinedl}
    \end{subfigure}
    \hfill
    \begin{subfigure}[t]{0.48\textwidth}
        \raggedleft
        \includegraphics[width=\textwidth]{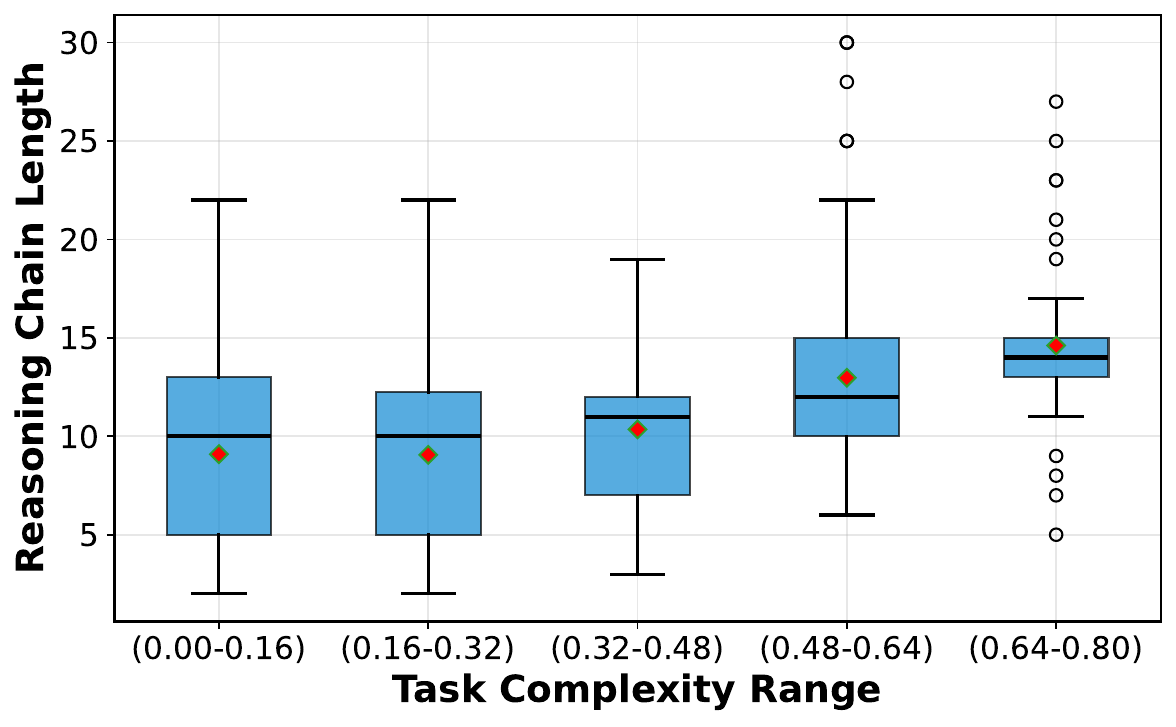}
        \caption{Tasks Complexity and reasoning chain length comparison. The lower complexity values indicate easier tasks.}
        \label{fig:combinedr}
    \end{subfigure}
    \caption{Analysis of chain length and task complexity.}
    \vspace{-0.5cm}
    \label{fig:combined}
\end{figure}

\subsection{Reasoning Length and Task Complexity}

To understand the robustness of {\selfanchor} to reasoning length and task difficulty, we conducted more detailed analyses. For each benchmark, we randomly sample 200 questions for analysis. Following prior work \cite{wu2025more,jin2024impact}, we quantify reasoning length by segmenting each model output into discrete reasoning steps and using the number of steps as a measure of reasoning depth. Task complexity is defined as $1 - \text{accuracy}$, where higher values indicate more challenging questions\footnote{Details on reasoning step segmentation and task complexity computation are provided in \autoref{appendix: Implementation and evaluation details}.}.

\paragraph{Robustness to Reasoning length.}
\label{sec:length}
We first examine the robustness of {\selfanchor} with respect to varying reasoning lengths. 
\autoref{fig:combinedl} illustrates the accuracy of {\selfanchor} compared to other baselines, across varying reasoning lengths.
{\selfanchor} achieves superior accuracy over the baselines across most reasoning lengths. While it consistently surpasses other methods, there is a slight performance dip relative to CoT on short sequences (length $\le 5$). This can be attributed to the design of {\selfanchor}, which is intended to mitigate attention drift in longer multi-step reasoning. Since simple tasks often can be solved in a few steps, they suffer less from attention drift.  
Moreover, {\selfanchor} exhibits stable and high accuracy even as the reasoning chain expands. 
These results demonstrate that {\selfanchor} is robust to reasoning length, effectively handling tasks regardless of whether they require short or extended reasoning chains.



\paragraph{Scaling Reasoning Length with Complexity.}
Next, we analyze how reasoning chain length scales with task difficulty. 
\autoref{fig:combinedr} shows the distribution of successful reasoning lengths across tasks of varying difficulty. 
The results show a clear trend that as complexity increases, {\selfanchor} tends to generate longer reasoning chains. 
To quantify this trend, we apply the Jonckheere-Terpstra test and obtain $\tau=0.102$ with $p=1.279\times10^{-7}$, indicating a statistically significant increasing trend across complexity levels. 
The effect size is modest, which is expected given variability in solution strategies and the large sample size. 
This aligns with the observation in \citet{wu2025more} that harder problems require longer reasoning chains to solve. We attribute this capability to the attention steering mechanism, which enables the model to focus on both problem context and immediate reasoning object throughout the reasoning, preventing attention drift as the reasoning chain extends.

In summary, our analysis highlights two takeaways: (1) {\selfanchor} demonstrates robustness across varying reasoning lengths. (2) \selfanchor\ encourage the model to generate longer reasoning chains for difficult problems, supporting its effectiveness in scaling to complex tasks.

\subsection{Failure Case Analysis}
\label{sec:fail}
To understand the failure modes in \selfanchor. We conducted a manual failure analysis on 300 randomly sampled failed cases (50 from each benchmark). Our analysis identified four primary failure modes:

\textbf{Reasoning Errors (42\%).} 
The most frequent failure mode involves the LLM making mistakes during the reasoning process. These include the misapplication of causal principles, flawed deductions, and incorrect conditional reasoning. 
For example, models sometimes exhibit causal misjudgment, where a condition that is necessary but not sufficient is incorrectly treated as the sole cause of an outcome.
We provide a concrete instance of this phenomenon from the BBH benchmark in Appendix~\ref{appendix:failure1}. Such errors suggest that, while attention steering helps maintain focus on relevant steps, it cannot fully compensate for weak logical priors or gaps in world knowledge.

\textbf{Misunderstanding the Problem (30\%).}
Another common failure mode stems from misunderstanding complex questions, leading to errors such as misidentifying all variables, misinterpreting the problem's requirements, or incorrectly parsing the relationships between entities. 
This issue is most common in multi-variable mathematical reasoning tasks, such as AQua.
We provide a representative example in Appendix~\ref{appendix:failure2}, where the model assigns wrong numerical values to the entity, causing the subsequent reasoning to an incorrect answer.

\textbf{Computational Errors (21.5\%).}
These failures include arithmetic mistakes, unit conversion errors, or algebraic slips. Even when the reasoning chain is correct, a single miscalculation often propagates to the final incorrect answer.

\textbf{Sensitivity to Plan Quality (6.5\%).}
As plan quality might affect downstream reasoning, we specifically examined the impact of planning quality within this category. We observed that only 6.5\% of total failures were attributable to planning errors. To determine if {\selfanchor} amplifies these errors, we cross-referenced these cases with one of the strongest baselines (RE2). We found that only 2.3\% of these specific cases were solvable by RE2. This suggests that when plans fail, the root cause is typically the intrinsic difficulty of the problem, not the anchoring mechanism. In other words, {\selfanchor} does not amplify errors. It reflects the model's existing understanding. Incorrect plans are strongly correlated with questions the base model already struggles with, regardless of attention steering.

\section{Related Work}
\textbf{LLM Reasoning.}
Prompt engineering has been widely adopted as a fundamental approach for enhancing LLM reasoning capabilities~\citep{liu2023pre,brown2020language}. Pioneered by Chain-of-Thought prompting~\citep{wei2022chain}, encourages explicit intermediate steps, which significantly improve performance on multi-step reasoning tasks. This has inspired numerous derivatives, from problem decomposition methods~\citep{wang2023plan,zhou2022least,khot2022decomposed,drozdov2022compositional}, to enhancing query comprehension~\citep{xu2024re,zheng2023take,mekala2023echoprompt,deng2023rephrase,mishra2022help}. 
While these methods demonstrate effectiveness in specific domains, they rely on predefined, static prompt for different tasks. On the other hand, LLMs remain sensitive to prompt variations and suffer from attention dilution during long generations~\citep{liu-etal-2024-lost,attention_decay, context_rot,lu2021fantastically,gu2024attention}. {\selfanchor} addresses these limitations by enabling the model to recalibrate its focus to the global objective defined by the original question and the local objective established by the current plan step at each stage of reasoning.


\textbf{Attention steering.} In contrast to the aforementioned prompt engineering, which devises better prompt strategies, attention steering methods directly guide LLMs during inference to emphasize the user-specified part of the context. 
Specifically, SPA \citep{tian2024selective} adjusts the logit probability distribution to emphasize the specified context. PASTA~\citep{zhang2023tell} identifies and reweights a subset of attention heads to redirect the model's attention to user-specified parts. 
SSA \citep{zhang2024selective} augments softmax with temperature scaling, whereas TOAST~\citep{shi2023toast} guides attention using learned feature selection.
However, these methods are static and task-specific, limiting their adaptability to diverse reasoning contexts. 
{\selfanchor} addresses this limitation by leveraging structured intermediate representations to enable dynamic attention steering.
\section{Conclusion}
We presented {\selfanchor}, a reasoning method that leverages the inherent structure of the reasoning chain to dynamically align attention during reasoning. Across six diverse reasoning benchmarks, {\selfanchor} consistently outperforms existing baselines. Notably, {\selfanchor} enhanced ``non-reasoning'' models achieve competitive performance compared with specialized reasoning models under most benchmarks. Moreover, our analysis reveals that \selfanchor's advantages are robust to varying task complexities. We hope {\selfanchor} serves as a step toward more reliable LLMs reasoning that requires neither parameter updates nor additional sampling.




\section*{Limitations}

Our method leverages the inherent structure of the reasoning chain to dynamically align attention during generation. However, as detailed in \autoref{sec:fail}, {\selfanchor} does not address intrinsic error modes of large language models. In particular, it cannot correct fundamental misunderstandings of the problem, nor can it reliably prevent arithmetic mistakes or algebraic slips. When such errors occur early, they may still propagate through subsequent steps despite improved attention alignment.

Furthermore, the headroom for inference-time intervention naturally diminishes when applying {\selfanchor} to larger or more powerful models, which suffer less from attention misalignment compared to smaller models. Nevertheless, {\selfanchor} still demonstrates performance gains on Qwen3-30B across all evaluated benchmarks, demonstrating its efficacy even at larger model scales.

\section*{Acknowledgments}
We thank the anonymous reviewers for their insightful feedback and the time and effort they devoted to reviewing our work. This work was supported in part by NSF grant CCF-2340408.

\bibliography{reference}

\appendix
\newpage

\newpage

\section{Alternative designs}
\label{appendix: alternative design}
\subsection{Attention Steering Strategies}

    As described in Section~\ref{Automatic Anchoring Strategy}, our primary {\selfanchor} design anchors attention to both the original question and the current plan step during reasoning generation. 
    To validate the design choices of {\selfanchor}, we evaluate three alternative anchoring strategies: anchoring solely to the question (\textit{Question-only}), solely to the current plan (\textit{Plan-only}), and to the full plan history (\textit{Plan history}). 
    
    We additionally compare against periodic summarization (\textit{Periodic Summarization}), a commonly used context-compression technique in multi-iteration agent trajectories~\citep{wu2025resum}. Since our setting uses single-shot reasoning (a single, uninterrupted chain of reasoning rather than multi-round interaction), we implemented a periodic summarization baseline where the model is prompted to compress its reasoning history into a concise summary state every $K$ reasoning steps. Results are presented in Table~\ref{tab:ablation_study}.
    


    \paragraph{\textit{Plan-only.}} A core hypothesis of {\selfanchor} is that anchoring only to the current sub-goal (plan step) is insufficient, because the model may lose sight of the global objective. Consistent with this intuition, \textit{Plan-only} substantially degrades performance, especially on the weaker Llama-3.1 model. This suggests that step-level guidance alone cannot reliably support long-horizon reasoning. 

 \begin{table*}[ht]
        \centering
        \setlength{\tabcolsep}{18pt}
            \centering
            \caption{Alternative design on anchoring strategies. Results show accuracy (\%) on AQuA-RAT and T4D benchmarks.}
            \begin{tabular}{ll|cc}
            \toprule
            \textbf{Model} & \textbf{Method}  & AQuA-RAT &  T4D \\
            \midrule
            \multirow{5.5}{*}{Llama3.1-8B} & CoT & 50.79 & 26.77   \\
                & {\selfanchor}  (Plan-only)& 40.94& 26.06\\
                & {\selfanchor} (Plan history)& 53.15& 36.70\\
                & Periodic summarization (K=2) & 46.06 & 39.82 \\
                \cmidrule(l){2-4}
            & {\selfanchor} (\textbf{Ours}) & \textbf{55.51} & \textbf{40.01} \\
            \midrule
            \multirow{5.5}{*}{Phi-4-mini} & CoT & 61.81 & 39.54\\
                & {\selfanchor} (Plan-only)& 50.00&47.34 \\
                & {\selfanchor} (Plan history) & 64.96 & 47.34\\
                & Periodic summarization (K=2) & 57.09 & 40.88 \\
                \cmidrule(l){2-4}
            & {\selfanchor} (\textbf{Ours}) & \textbf{68.50} & \textbf{49.47} \\
            \bottomrule
            \end{tabular}
            \label{tab:ablation_study}
    \end{table*}
    
    \paragraph{Impact of History.} 
    We also evaluate an alternative design that anchors to all prior plan steps in addition to the current step and the original question. The motivation is that preserving attention to the full reasoning trajectory might provide helpful historical context. Specifically, we compare two anchoring strategies:
    \begin{itemize}
    \item \textbf{{\selfanchor} (Ours):} Anchors to the original question and current plan step only, where $a_i = \{\text{Question}, \text{Plan}_i\}$
    \item \textbf{Plan history:} Anchors to the original question, current plan step, and all prior plan steps, where  $a_i = \{\text{Question}, \text{Plan}_1, \text{Plan}_2, \ldots, \text{Plan}_i\}$
    \end{itemize}
    
    Anchoring to the entire \textit{Plan history} also yields suboptimal results. It is likely because anchoring to all plan history will dilute attention across too many past steps, weakening focus on the immediate reasoning objective.
    
    \paragraph{Periodic Summarization.}
    The result shows that {\selfanchor} consistently outperforms the periodic summarization method across all benchmarks and models.
    This is likely because summary compression introduces information loss and can overwrite fine-grained constraints that remain useful for downstream steps.

    In conclusion, our proposed design achieves the best performance. These results support our central design principle: effective reasoning-time attention alignment requires maintaining both global task awareness and step-wise local focus.

    \begin{table*}[ht]
        \centering
        \setlength{\tabcolsep}{14pt}
            \centering
            \caption{Alternative design on confidence aggregation methods. ``Fixed Strength" uses a constant anchor weight, while others calculate dynamic weights based on the mean token probability. Results show accuracy (\%).}
            \begin{tabular}{ll|cc}
            \toprule
            \textbf{Model} & \textbf{Method}  & AQuA-RAT &  T4D \\
            \midrule
            \multirow{5.5}{*}{Llama3.1-8B} & CoT & 50.79 & 26.77   \\
                & {\selfanchor} (Fixed Strength) & 47.64 & 36.17 \\
                \cmidrule(l){2-4}
                & {\selfanchor} (Entropy) & 47.64 & 35.64 \\
                \cmidrule(l){2-4}
                & {\selfanchor} (Arithmetic) & 54.72 & 35.99 \\
                & {\selfanchor} (Geometric) & 55.11 & 35.28 \\
                & \textbf{{\selfanchor} (Harmonic)} & \textbf{55.51} & \textbf{40.01}  \\
            \midrule
            \multirow{5.5}{*}{Phi-4-mini} & CoT & 61.81 & 39.54\\
                & {\selfanchor} (Fixed Strength) & 62.60 & 48.58 \\
                \cmidrule(l){2-4}
                & {\selfanchor} (Entropy) & 60.63 &  49.30\\
                \cmidrule(l){2-4}            
                & {\selfanchor} (Arithmetic) & 67.71 & 48.40 \\
                & {\selfanchor} (Geometric) & 67.72 & 49.11 \\
                & \textbf{{\selfanchor} (Harmonic)} & \textbf{68.50}  & \textbf{49.47}\\
            \bottomrule
            \end{tabular}
            \label{tab:means_selection}
    \end{table*}

    \subsection{Dynamic Attention Alignment Strength}
    \label{appendix:mean}
    We evaluate different methods for calculating confidence scores by aggregating token-level probabilities into a step-level confidence score. Let $P_i = \{p_1, p_2, \ldots, p_m\}$ represent the set of token-level confidence scores for tokens generated in the current reasoning step $i$. We compare three approaches for calculating average confidence scores against a baseline of \textbf{Fixed Anchor Strength} (where the dynamic weight is replaced by a constant scalar).

    The aggregation methods are defined as follows:
    
    \textbf{Harmonic Mean:}
    \begin{equation}
    p_{\text{harmonic}} = \frac{n}{\sum_{i=1}^{n} \frac{1}{p_i}}
    \end{equation}
    
    \textbf{Geometric Mean:}
    \begin{equation}
    p_{\text{geometric}} = \left(\prod_{i=1}^{n} p_i\right)^{1/n}
    \end{equation}
    
    \textbf{Arithmetic Mean:}
    \begin{equation}
    p_{\text{arithmetic}} = \frac{1}{n}\sum_{i=1}^{n} p_i
    \end{equation}

    Lastly, we consider the normalized predictive \textbf{Entropy} of the model's next-token distribution. 
    
    
    
    Table~\ref{tab:means_selection} presents the results across two reasoning benchmarks using Llama3.1 and Phi-4-mini models. 
    First, we observe that the Fixed Anchor Strength strategy significantly underperforms compared to dynamic methods, confirming that adaptive anchoring based on model confidence is essential for effective reasoning. 
    Among the dynamic methods, the Harmonic Mean consistently outperforms both geometric and arithmetic means across all settings. This is likely because the harmonic mean is heavily penalized by minimum values. In reasoning tasks, a single low-confidence token signals potential attention drift. 
    The geometric mean performs second-best, as it also penalizes low values more than the arithmetic mean. The arithmetic mean shows the weakest performance, as it can be dominated by high-confidence tokens and may miss instances where attention drift occurs for specific reasoning components.
    Lastly, the entropy variant does not improve over the harmonic mean and can slightly degrade performance, suggesting that entropy introduces additional noise under our setting and is less aligned with the specific failure patterns that anchoring aims to correct.

\section{Efficiency}
While {\selfanchor} outperforms existing methods, we further analyze its efficiency. We use the second per task as the metric. As shown in \autoref{tab: efficiency}, ToT incurs approximately $4\times$ higher latency than {\selfanchor}. The high latency of ToT primarily arises from the tree search procedure, which requires generating and evaluating multiple candidate reasoning branches at each step. In contrast, {\selfanchor} is modest compared to existing reasoning methods, while delivering substantially higher accuracy. 
Specifically, to steer attention to specific tokens, it performs an additional masked forward pass to compute logits reweighting. 

\begin{table*}[ht]
    \scriptsize
    \centering
    \setlength{\tabcolsep}{4pt}
    \caption{Efficiency comparison}
    \resizebox{\textwidth}{!}{
        \begin{tabular}{l|ccccc}
    \toprule
    \multirow{2.5}{*}{\textbf{Model}} &  \multicolumn{5}{c}{Latency (Second/Task) $\downarrow$} \\
    \cmidrule(lr){2-6}
    & CoT & PS+ & RE2 & ToT & \textbf{{\selfanchor}} \\
    \midrule
    \multirow{1}{*}{Llama3.2-3B} & 34.96 & 35.20 (+0.24) & 27.64 (-7.32) & 193.55 (+158.59) & 48.06 (+13.10) \\
    \midrule
    \multirow{1}{*}{Phi-4-mini-4B} & 26.26 & 21.58 (-4.68) & 24.27 (-1.99) & 278.70 (+252.44) & 42.30 (+16.04) \\
    \midrule
    \multirow{1}{*}{Qwen3-4B} & 61.24 & 89.97 (+28.73) & 50.40 (-10.84) & 292.28 (+231.04) & 74.29 (+13.05)\\
    \midrule
    \multirow{1}{*}{Llama3.1-8B} & 109.82 & 110.54 (+0.72) & 101.31 (-8.51) & 428.2 (+318.38) & 89.71 (-20.11)\\
    \midrule
    \multirow{1}{*}{Phi-4-15B} & 108.44 & 108.69 (+0.25) & 108.66 (+0.22)& 482.48 (+374.04) & 122.59 (+14.15)\\
    \midrule
    \multirow{1}{*}{Qwen3-30B} & 468.99 & 536.02 (+67.03) &  424.9 (-44.09) & 1790.36 (+1321.37) & 459.03 (-9.96)\\
    \bottomrule
    \end{tabular}
    }
    \label{tab: efficiency}
\end{table*}

\section{Related work: Reasoning LLMs}
Recent researches demonstrate specialized training pipelines can internalizing reasoning capabilities directly into the model. 
State-of-the-art models, such as OpenAI o1~\citep{gpt_o1} and DeepSeek-R1~\citep{deepseekai2025deepseekr1incentivizingreasoningcapability}, leverage fine-tuning and large-scale reinforcement learning~\citep{RL1,RL2} on reasoning trajectories, achieving remarkable performance on complex reasoning tasks. 
Despite their impressive performance, they require substantial computation and a considerable amount of training data. 
More critically, even powerful reasoning models remain vulnerable to attention dilution during extended generation~\citep{liu-etal-2024-lost,gu2024attention,xiao2023efficient}, where the model fails to sustain focus on relevant context. 
In comparison, {\selfanchor} provides a training-free reasoning method that uses the inherent structure of the reasoning chain to align attention during generation. Our results show that {\selfanchor} enables ``non-reasoning'' models reach better or comparable performance compared to specialized reasoning models of the same size and family across most settings.

\section{\selfanchor\ Prompt Details}
\autoref{tab:selfanchorprompt} presents the prompt used by {\selfanchor} to elicit explicit planning and step-by-step reasoning in a structured format. The prompt instructs the model to first decompose the original problem into a sequence of plan steps, followed by corresponding actions and reasoning for each step, ensuring that intermediate reasoning remains aligned with the overall problem specification.

\begin{table*}[ht]
  \centering
  \small
  \begin{tabular}{>{\raggedright\arraybackslash\ttfamily}p{0.95\textwidth}<{}}
      \toprule
      \headercolorlong
        \textbf{TASK INSTRUCTIONS}\\
          You are an expert problem solver. Your task is to decompose the given problem into a clear, step-by-step plan, reasoning the plan, and solve the problem step by step in JSON format.\\
            For each plan step, provide a key-value pair: the key is the plan step (as stated), the value is the detailed reasoning and action for that step.\\
            Now, implement a reasoning structure to follow step-by-step and arrive at correct answers in JSON format.\\\\
        \headercolorlong
        \textbf{TASK}\\
            Original problem: {\textbf{\{question\}}}\\
      \bottomrule
  \end{tabular}
  \caption{Prompt used by \selfanchor\ to elicit structured planning and step-wise reasoning in format.}
  \label{tab:selfanchorprompt}
\end{table*}

\section{Implementation and evaluation details}
    \label{appendix: Implementation and evaluation details}
    \subsection{Prompt example}
    \label{appendix:prompt}

        \paragraph{Chain-of-Thought and Re-Reading.}
        \autoref{tab:cotprompt} presents the prompts used by the Chain-of-Thought and Re-Reading. The Chain-of-Thought prompt encourages step-by-step reasoning, while the Re-Reading prompt reinforces attention to the original question by explicitly repeating it before reasoning.
        
        \begin{table}[ht]
          \centering
          \small
          \begin{tabular}{>{\raggedright\arraybackslash\ttfamily}p{0.45\textwidth}<{}}
              \toprule
              \headercolorlong
                \textbf{Chain-of-Thought}\\
                  Let's think step by step.\\\\
                  \headercolorlong
                  \textbf{Re-Reading.}\\
                  {\textbf{\{question\}}}\\
                  Read the question again:\\
                {\textbf{\{question\}}}\\
              \bottomrule
          \end{tabular}
          \caption{Prompt used by Chain-of-Thought and Re-Reading.}
          \label{tab:cotprompt}
        \end{table}

        \paragraph{Plan-and-solve+.}
        We adopt the implementation of Plan-and-Solve+~\citep{wang2023plan}. For mathematical reasoning tasks, we use a prompt that explicitly emphasizes variable extraction, planning, and step-by-step numerical reasoning. For non-mathematical (commonsense) tasks, we adopt a simplified planning-based prompt, as shown in \autoref{tab:planandsolveprompt}.

        \begin{table*}[ht]
          \centering
          \small
          \begin{tabular}{>{\raggedright\arraybackslash\ttfamily}p{0.95\textwidth}<{}}
              \toprule
              \headercolorlong
                \textbf{Mathematical Reasoning Tasks}\\
                  Let's first understand the problem, extract relevant variables and their corresponding numerals, and make and devise a complete plan. Then, let's carry out the plan, calculate intermediate variables (pay attention to correct numerical calculation and commonsense), solve the problem step by step, and show the answer.\\\\
        
                \headercolorlong
                \textbf{COMMONSENSE TASKS}\\
                  Let's first prepare relevant information and make a plan. Then, let's answer the question step by step (pay attention to commonsense and logical coherence).\\
              \bottomrule
          \end{tabular}
          \caption{Prompts used for the Plan-and-Solve+ baseline under different task settings.}
          \label{tab:planandsolveprompt}
        \end{table*}

    \subsection{Evaluation Details}
     Our experiments were conducted on the NVIDIA Quadro RTX 8000 GPU. To improve inference efficiency, we employ quantized inference, specifically NF4 4-bit quantization with FP16 computation. We observe some performance drop compared to full-precision inference, which is consistent with prior studies showing that quantization can negatively affect performance, particularly for smaller models and long reasoning generations~\citep{mekala2025does,liu2025quantization,li2025quantization}. 
     
    To verify that our conclusions remain robust under full-precision inference, we further conduct a verification experiment on MATH-500 under full precision. The results show that {\selfanchor} consistently outperforms CoT across both models: Llama3.1 improves from 46.8 to 52.2, and Qwen3-4B improves from 81.6 to 90.4. This suggests that while quantization may affect absolute performance, the relative advantage of {\selfanchor} remains stable.
    
    During generation, we use greedy decoding. We adopt standard metrics used in prior work \citep{chuang2024dola,wang2023plan,zhou2024self}, including accuracy and exact match, for AQuA, BBH, T4D, and MATH. For GSM8K and StrategyQA, we follow the factual accuracy evaluation protocol introduced by \citet{chuang2024dola}.For the analysis presented in \autoref{sec:length}, we examine model performance across different reasoning chain lengths on Llama3.2-4B.

    To ensure consistent answer extraction, we prompt all models to conclude their response with the phrase: ``Conclude with the final answer using the format: ``Final answer": "$<$your answer$>$'' where $<$your answer$>$ denotes either a multiple-choice option or a string answer. We then apply task-specific heuristics to extract $<$your answer$>$ from the output.

    \subsection{Implementation Details}
    We follow the definitions of reasoning step segmentation and task complexity from \citep{wu2025more,jin2024impact}. The performance gain of a task is computed as the difference in accuracy between two methods on that task. 
    
    \paragraph{Reasoning Step Segmentation.}
    To quantify reasoning length, we measure the number of discrete reasoning steps in each model-generated result. Rather than using token count, which can be noisy and model-dependent, we adopt step count as a more meaningful indicator of reasoning depth. Specifically, we segment the full reasoning output by newline characters (``\textbackslash n''), remove empty lines resulting from consecutive newlines (``\textbackslash n\textbackslash n''), and treat each remaining line as one reasoning step. The total number of such lines is taken as the reasoning length for that example.
    
    For datasets with high variability in reasoning length (e.g., MATH-500), we normalize step counts using fixed-width bins to ensure sufficient samples per group. We use a bin width of 5 when comparing reasoning length distributions across models, and a finer bin width of 2 when analyzing the relationship between reasoning length and task complexity. We verify that using smaller bin widths yields qualitatively similar results.
    
    \paragraph{Task Complexity.}
    Task complexity is defined based on empirical difficulty. For each question, we compute complexity as $1 - \text{accuracy}$, where accuracy is averaged over all evaluated models. Higher values correspond to more challenging questions.
    

\section{Case Study}
\label{appendix:case}
    In this section, we provided qualitative examples to illustrate the reasoning trajectory generated by each method. 

    \subsection{Failure Case 1}
    \label{appendix:failure1}

    Table\ref{tab:fail1} demonstrates the failure to distinguish between a necessary background condition and a causal difference-maker. While the model correctly identifies the black wire as a necessary condition (Step 4), it fails to apply the requisite world knowledge regarding causal selection. Specifically, the distinction between a standing ``condition'' and a precipitating ``cause''. In Step 5, the system explicitly articulates a flawed logical prior, asserting that any necessary but insufficient factor qualifies as a cause. Consequently, the model ignores the principle of normality, attributing causality to the component functioning within its design parameters (the black wire) rather than the deviating factor (the red wire), leading to an invalid conclusion.

    \begin{table*}
    \centering
    \caption{An example of BBH causal reasoning. Incorrect causal inference is highlighted in \lightred{red}.}
    \resizebox{\linewidth}{!}{
    \begin{tabular}{l l}
        \toprule
        \multirow{3}{*}{\textbf{Prompt}}
        & \textit{\textbf{Question:}} A machine is set up in such a way that it will short circuit if both the black wire and the red wire touch the battery\\
        & at the same time. The machine will not short-circuit if just one of these wires touches the battery. The black wire is designated\\
        &  as the one that is supposed to touch the battery, while the red wire is supposed to remain in some other part of the machine. \\
        & One day, the black wire and the red wire both end up touching the battery at the same time. There is a short circuit. Did the \\
        & black wire cause the short circuit? \\
        & \textit{\textbf{Options: (A) Yes \quad (B) No}} \\
        \midrule
        \textbf{\selfanchor} & \{  "Step 1: Understand the scenario": "The machine is designed to short circuit only when both the black wire \\
        & and the red wire touch the battery at the same time.",\\
        & "Step 2: Identify the necessary condition for the short circuit": "The necessary condition for the short circuit is \\
        & that both the black wire and the red wire touch the battery at the same time.", \\
        & "Step 3: Determine if the black wire's action was sufficient to cause the short circuit": "The black wire's action alone is not \\
        & sufficient to cause the short circuit, as the machine will not short circuit if just one of these wires touches the battery.", \\
        & "Step 4: Determine if the black wire's action was necessary to cause the short circuit": "The black wire's action was necessary to\\
        & cause the short circuit, as the machine will not short circuit if just the red wire touches the battery.", \\
        & "Step 5: \lightred{Apply the concept of causation": "In causation, if the action of one thing is necessary for the occurrence of an event,} \\
        & \lightred{but not sufficient on its own, then that thing is considered to have caused the event.",} \\
        & "Step 6: Determine if the black wire caused the short circuit": "Based on the necessary condition for the short circuit and the \\
        & concept of causation, the black wire did cause the short circuit.",\\
        & "Final answer": "\lightred{(A) Yes}" \} \\
        \bottomrule
    \end{tabular}
    }
    \label{tab:fail1}
\end{table*}

    \subsection{Failure Case 2}
    \label{appendix:failure2}
    Table\ref{tab:fail2} illustrates a representative instance of entity-attribute binding failure. 
    The prompt specifies a relative increase in purchasing power (``5 more articles''). However, the model interpreted ``5 more'' as a fixed quantity of ``6'' (likely assuming she bought 1 before). This failure to distinguish between a marginal change and a total count results in the formulation of a fallacious linear equation, bypassing the necessary comparative logic required to solve for the original price.
    
    \begin{table*}
    \centering
    \caption{An example of mathematical reasoning in Aqua. Incorrect numerical reasoning is highlighted in \lightred{red}.}
    \resizebox{\linewidth}{!}{
    \begin{tabular}{l l}
        \toprule
        \multirow{4}{*}{\textbf{Prompt}}
        & \textit{\textbf{Question:}} Anna is able to buy 5 more articles for \$300 after the price of each article decreased by 15\%. \\
        & What is the new selling price of each article? \\
        & \textit{\textbf{Options: (A) \$8 \quad (B) \$10 \quad (C) \$13.6 \quad (D) \$22.9 \quad (E) \$40}} \\
        \midrule
        \textbf{\selfanchor}
        & \{ "Step 1: Establish variables": "Let the original price of each article be \( x \).", \\
        & "Step 2: Apply price decrease": "After a 15\% decrease, the new price becomes \( 0.85x \).", \\
        & "Step 3: Set up the cost equation": "Anna buys 5 more articles for \$300, so the total cost is represented as \\
        &  \lightred{6} \(\times 0.85x = 300 \).", \\
        & "Step 4: Solve for the original price": "Solving the equation gives \( x = \frac{300}{6 \times 0.85} = 58.82 \).", \\
        & "Step 5: Compute the new selling price": "The new price is \( 0.85 \times 58.82 = 50 \).", \\
        & "Step 6: Match with answer choices": "The value \$50.00 does not appear in the options, \\
        & leading to an incorrect selection based on proximity.", \\
        & "Final answer": "\lightred{(D) \$22.9}" \} \\
        \bottomrule
    \end{tabular}
    }
    \label{tab:fail2}
\end{table*}

    \subsection{Comparison Example 1}
    \autoref{tab:An example of anchor} illustrates a representative example where attention steering prevents reasoning errors. In Steps 2 and 4, the {\selfanchor} without attention steering approach incorrectly identifies ``her" as referring to ``the secretary"; this may be because models over-focus on sentence subjects rather than maintaining focus on the syntactic relationships that determine pronoun reference. In contrast, {\selfanchor} correctly identifies that ``her" refers to ``the analyst" by maintaining attention on both the original question context and the current reasoning step.

    \begin{table*}
    \centering
    \caption{An example of BBH. Incorrect inferences are highlighted in \lightred{red}. Correct ones are highlighted in \lightgreen{green}.}
    \resizebox{\linewidth}{!}{
    \begin{tabular}{l l}
        \toprule
        \multirow{2}{*}{\textbf{Prompt}}  
        & \textit{\textbf{Sentence: The secretary came to the analyst's office and helped her to book a flight.}} \\
        & \textit{\textbf{Options: (A) Helped the secretary book a flight (B) Helped the analyst book a flight (C) Ambiguous}} \\
        \midrule
        \textbf{\selfanchor} &  \{  "Step 1: Identify the pronoun": "The pronoun is 'her'.",\\
         \text{(w/o attention steering)} &     "Step 2: Determine the antecedent of the pronoun": "\lightred{The antecedent of 'her' is the subject of the sentence, which is 'The secretary'}", \\
         \textbf{} & "Step 3: Analyze the sentence structure": The subject of the sentence is 'The secretary', and the verb is 'came'. The object of the \\
         & \:\:\:\: verb 'came' is 'to the analyst's office'. The verb 'helped' has the pronoun 'her' as its object.",\\
         & "Step 4: Determine the antecedent of the pronoun based on the sentence structure": "Based on the sentence structure, the antecedent \\
         & \:\:\:\: of the pronoun \lightred{'her' is the subject of the sentence, which is 'The secretary'}.",\\
         & "Step 5: Choose the correct option": "The correct option is \lightred{(A) Helped the secretary book a flight.}"\}\\
        \midrule
        \textbf{{\selfanchor}} &  \{  "Step 1: Identify the pronoun": "The pronoun in the sentence is 'her'.",\\
         &   "Step 2: Determine the antecedent of the pronoun": "To find the antecedent, we need to look for the noun that 'her' refers to. In this\\
         &  \:\:\:\: case, \lightgreen{'her' is a possessive pronoun that refers to the analyst.}", \\
         & "Step 3: Analyze the sentence structure": "The subject of the sentence is 'The secretary', and the object of the verb 'helped' is 'her'",\\
         & "Step 4: Identify the antecedent based on the sentence structure": "\lightgreen{'her' refers to the analyst, not the secretary.}",\\
         & "Step 5: Choose the correct option": "Since 'her' refers to the analyst, the correct option is \lightgreen{(B) Helped the analyst book a flight.}" \}\\
        \bottomrule
    \end{tabular}
    }
    \label{tab:An example of anchor}
    \end{table*}

    \subsection{Comparison Example 2}
    In this example, both CoT (\autoref{tab:example1_1}) and RE2 (\autoref{tab:example3_1}) know English is the official language, but they drift into lengthy cultural descriptions and ultimately provide the wrong polarity. PS+ (\autoref{tab:example2_1}) follows a reasonable reasoning path, but the final answer again diverges from the original question. {\selfanchor} (\autoref{tab:example4_1}) maintains focus on the question and produces the correct answer.

\begin{table*}
    \centering
    \caption{An example illustrating Chain-of-Thought's reasoning process.}
    \resizebox{\linewidth}{!}{
    \begin{tabular}{l l}
        \toprule
        \textbf{Question:} & Is the language used in Saint Vincent and the Grenadines rooted in English?\\
        \midrule
        \textbf{Chain-of-Thought (CoT)} &  No, the official language of Saint Vincent and the Grenadines is English, but the local Creole language, \\
        & Saint Vincentian Creole, is also widely spoken. Saint Vincentian Creole is a English-based Creole language that \\
        & incorporates elements of French, and African languages. It is the mother tongue of the majority of the population\\
        & and is used in everyday conversation, music, and literature. While English is the official language and is used in \\
        & government, education, and formal settings, Saint Vincentian Creole is the language of the people and is an \\
        & important part of the country's cultural identity. So, while English is the official language, the local Creole \\
        & language is also deeply rooted in the culture and daily life of Saint Vincent and the Grenadines.\\
         & Final answer: No\\
        \bottomrule
    \end{tabular}
    }
    \label{tab:example1_1}
    \end{table*}

\begin{table*}
    \centering
    \caption{An example illustrating Plan-and-Solve's reasoning process.}
    \resizebox{\linewidth}{!}{
    \begin{tabular}{l l}
        \toprule
        \textbf{Question:} & Is the language used in Saint Vincent and the Grenadines rooted in English?\\
        \midrule
        \textbf{Plan-and-Solve (PS+)} &  Step 1: Understand the question \\
        & The question asks about the language used in Saint Vincent and the Grenadines being rooted in English. \\
        & This means we need to determine the primary language spoken in the country and its linguistic origins.\\
        & Step 2: Research \\
        &  To answer this question, we need to research the languages spoken in Saint Vincent and the Grenadines. Saint \\
        &  Vincent and the Grenadines is an island nation in the southern part of the Windward Islands in the Caribbean Sea. \\
        &  Step 3: Analyze the findings\\
        &  After researching, we find that the official language of Saint Vincent and the Grenadines is English. However, the\\
        & country also has a Creole language known as Vincentian Creole, which is a mix of English, French, and African languages.\\
        & Step 4: Formulate the answer \\
        & Given that the primary language used in Saint Vincent and the Grenadines is English, but the Creole language is also \\
        & widely spoken and has roots in English, French, and African languages, the language used in Saint Vincent and the \\
        & Grenadines is rooted in English to a significant extent.\\
        & Step 5: Provide the final answer \\
        &  The answer is English, with Vincentian Creole, a mix of English, French, and African languages, also being widely spoken.\\
        & Final answer: The final answer is English, with Vincentian Creole also being a significant part of the language \\
        & landscape in Saint Vincent and the Grenadines.\\
        \bottomrule
    \end{tabular}
    }
    \label{tab:example2_1}
    \end{table*}

\begin{table*}
    \centering
    \caption{An example illustrating Re-Reading's reasoning process. }
    \resizebox{\linewidth}{!}{
    \begin{tabular}{l l}
        \toprule
        \textbf{Question:} & Is the language used in Saint Vincent and the Grenadines rooted in English?\\
        \midrule
        \textbf{Re-Reading (RE2)} &  No, the language used in Saint Vincent and the Grenadines is primarily English, but it also includes \\
        & elements of French, African, and Indian languages due to its colonial history and diverse population. \\
        & Final answer: No\\
        \bottomrule
    \end{tabular}
    }
    \label{tab:example3_1}
    \end{table*}

\begin{table*}
    \centering
    \caption{An example illustrating {\selfanchor}'s reasoning process.}
    \resizebox{\linewidth}{!}{
    \begin{tabular}{l l}
        \toprule
        \textbf{Question:} & Is the language used in Saint Vincent and the Grenadines rooted in English?\\
        \midrule
        \textbf{{\selfanchor}} &  \{"Understand the question.": "\\
        & The question is asking whether the language spoken in Saint Vincent and the Grenadines is based on English.",\\
        & "Research the official language of Saint Vincent and the Grenadines.": \\
        &"Saint Vincent and the Grenadines is an island nation in the Caribbean, and its official language is English.",\\
        &"Confirm the linguistic roots of the official language.": \\
        &"English is a Germanic language, which means it has roots in Old English, a language spoken in parts of what is now England."\}\\
        & Final answer: Yes\\
        \bottomrule
    \end{tabular}
    }
    \label{tab:example4_1}
    \end{table*}

\section{Numerical Experiment Result on Big-Bench-Hard}
    \label{appendix:bbh}

\begin{table*}[t]
    
    \centering
    \setlength{\tabcolsep}{2pt}
    \caption{BBH task categories and the number of correct cases in each category.}
    \resizebox{1\textwidth}{!}{
    \begin{tabular}{llcccccc}
        \toprule
        \multirow{2}{*}{\textbf{Model}} & \multirow{2}{*}{\textbf{Method}} & \multicolumn{5}{c}{\textbf{BBH}} \\
        \cmidrule(lr){3-8}
        & & disambiguation\_qa & causal\_judgement  & date\_understanding & logical\_deduction & salient\_translation\_error\_detection & snarks\\
        \midrule
        \multirow{5}{*}{Llama3.2-3B} & CoT & 88 & 71 & 102 & 34 & 95 & 89\\
         & PS+ & 104 & 63 & 97 & 86 & 98 & 105 \\
         & re-read & 74 & 84 & 127 & 40 & 89 & 105\\
         & ToT & 81 & 90 & 120 & 73 & 93 & 107 \\
        & {\selfanchor} & 131 & 101 & 131 & 101 & 96 & 129\\
        \midrule
        \multirow{5}{*}{Phi4-mini-4B} & CoT & 162 & 113 & 154 & 114 & 149 & 134\\
         & PS+ & 160 & 126 & 138 & 130 & 136 & 124\\
         & re-read & 162 & 115 & 165 & 113 & 135 & 148\\
         & ToT & 151 & 102 & 129 & 114 &  96 & 114 \\
        & {\selfanchor} & 152 & 116 & 174 & 128 & 140 & 142\\
        \midrule
        \multirow{5}{*}{Qwen3-4B} & CoT & 183 & 121 & 174 & 208 & 167 & 148\\
         & PS+ & 179 & 108 & 165 & 207 & 160 & 150 \\
         & re-read & 179 & 121 & 203 & 230 & 147 & 154\\
         & ToT & 172 & 114 & 176 & 219 & 122 & 150 \\
        & {\selfanchor} & 185 & 115 & 193 & 224 & 161 & 150\\
        \midrule
        \multirow{5}{*}{Llama3.1-8B} & CoT & 100 & 91 & 182 & 67 & 110 & 125\\
         & PS+ & 138 & 108 & 136 & 78 & 124 & 122 \\
         & re-read & 124 & 98 & 190 & 85 & 121 & 127\\
         & ToT & 114 & 104 & 153 & 87 & 120 & 111 \\
        & {\selfanchor} & 152 & 100 & 160 & 128 & 132 & 127\\
        \midrule
        \multirow{5}{*}{Phi-4-15B} & CoT & 180 & 120 & 203 & 169 & 152 & 160\\
         & PS+ & 180 & 117 & 140 & 186 & 155 & 156\\
         & re-read & 179 & 120 & 221 & 191 & 158 & 154 \\
         & ToT & 145 & 90 & 198 & 147 &  145 & 150\\
        & {\selfanchor} & 176 & 122 & 218 & 203 & 148 & 161 \\
        \midrule
        \multirow{5}{*}{Qwen3-30B} & CoT & 175 & 125 & 182 & 185 & 172 & 155 \\
         & PS+ & 184 & 122 & 172 & 181 & 165 & 156\\
         & re-read & 183 & 125 & 185 & 210 & 174 & 156\\
         & ToT & 180 & 132 & 195 & 186 & 158 & 153 \\
        & {\selfanchor} & 180 & 123 & 196 & 200 & 160 & 157\\
        \bottomrule
    \end{tabular}}
    \label{tab:truth}
\end{table*}

\end{document}